\documentclass[conference]{IEEEtran}
\IEEEoverridecommandlockouts
\usepackage{cite}
\usepackage{amsmath,amssymb,amsfonts}
\usepackage{algorithmic}
\usepackage{graphicx}
\usepackage{textcomp}
\usepackage{xcolor}
\usepackage{booktabs}
\usepackage{amsmath}
\usepackage{xcolor}
\usepackage{colortbl} 
\usepackage{booktabs} 
\usepackage{array}    
\usepackage{float}
\usepackage[colorlinks=false, hidelinks]{hyperref}
\usepackage{graphicx}  
\def\BibTeX{{\rm B\kern-.05em{\sc i\kern-.025em b}\kern-.08em
    T\kern-.1667em\lower.7ex\hbox{E}\kern-.125emX}}

\begin{document}
\title{CDFormer: Cross-Domain Few-Shot Object Detection Transformer Against Feature Confusion}



\author{Boyuan Meng, Xiaohan Zhang, Peilin Li, Zhe Wu, Yiming Li, Wenkai Zhao, Beinan Yu, \\and Hui-Liang Shen, \emph{Senior Member, IEEE}

\thanks{	
	Boyuan Meng, Xiaohan Zhang, Peilin Li, Zhe Wu, Yiming Li, Wenkai Zhao, and Hui-Liang Shen are with the College of Information Science and Electronic Engineering, Zhejiang University, Hangzhou 310027, China (e-mail: \{mengby, zhangxh2023, peilinli, jeffw, yimingl.20, zwkzju, shenhl\}@zju.edu.cn).
		
	Beinan Yu is with the College of Computer Science and Technology, Zhejiang University, Hangzhou 310027, China, and also with the Jinhua Institute of Zhejiang University, Jinhua 321299, China (e-mail: mr\_vernon@hotmail.com).
}}

\maketitle

\begin{abstract}
Cross-domain few-shot object detection (CD-FSOD) aims to detect novel objects across different domains with limited class instances. Feature confusion, including object-background confusion and object-object confusion, presents significant challenges in both cross-domain and few-shot settings. In this work, we introduce CDFormer, a cross-domain few-shot object detection transformer against feature confusion, to address these challenges. The method specifically tackles feature confusion through two key modules: object-background distinguishing (OBD) and object-object distinguishing (OOD). The OBD module leverages a learnable background token to differentiate between objects and background, while the OOD module enhances the distinction between objects of different classes. Experimental results demonstrate that CDFormer outperforms previous state-of-the-art approaches, achieving 12.9\% mAP, 11.0\% mAP, and 10.4\% mAP improvements under the 1/5/10 shot settings, respectively, when fine-tuned. Code is available at \textcolor{blue}{\url{https://longxuanx.github.io/CDFormer/}}.

\end{abstract}


\section{Introduction}
\label{sec:intro}

Few-shot object detection (FSOD) has been widely adopted in handling situations with limited data. It is trained on the base classes with sufficient data and detects expected objects of novel classes based on a few examples. The current FSOD approaches, which can be categorized to meta-learning-based ones \cite{DE-ViT,Meta-RCNN,FCT,Meta-YOLO,Meta-DETR,FS-DETR} and fine-tuning-based ones \cite{TFA,DeFRCN,FSCE,Retentive_RCNN}, have achieved remarkable performance on FSOD benchmarks. 

\begin{figure}[!htb]
  \centering
  \includegraphics[width=0.85\linewidth]{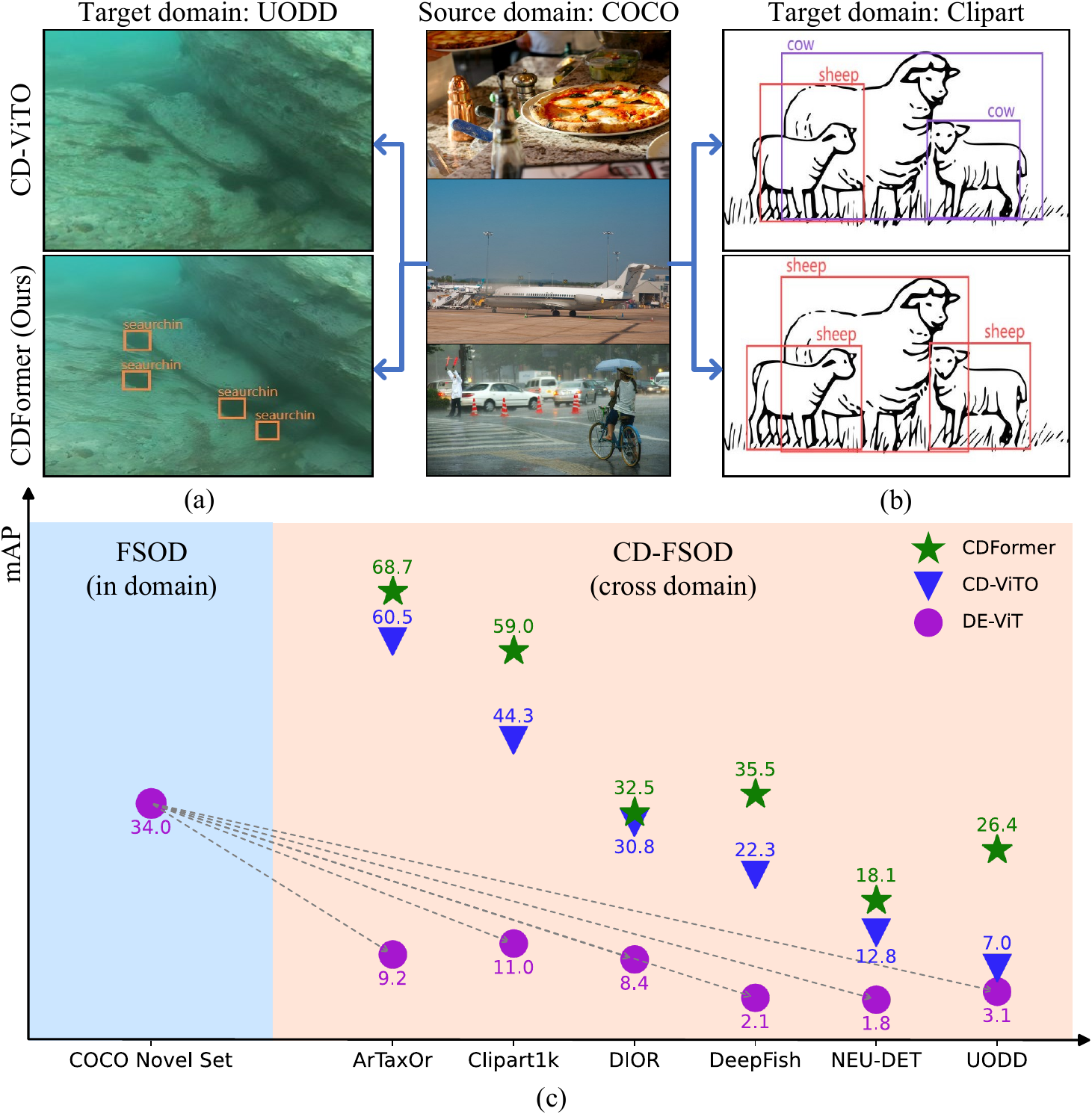}
   \caption{Illustration of the task, challenges, and our performance of the CD-FSOD. (a) Object-background confusion. (b) Object-object confusion. (c) Performance comparison between our CDFormer and the state-of-the-art DE-ViT \cite{DE-ViT} and CD-ViTO  \cite{CD-ViTO}.}
   \label{fig:intro}
\vspace{-1.5em}
\end{figure}

The current FSOD works usually assume that the base classes and novel classes belong to the same domain. However, real applications often require cross-domain few-shot object detection (CD-FSOD) \cite{CD-ViTO}. For example, as shown in Fig. \ref{fig:intro}(a) and Fig. \ref{fig:intro}(b), the base classes consist of everyday real-world scenes, while the novel classes come from cartoon or underwater scenarios. Consequently, even the advanced FSOD work DE-ViT \cite{DE-ViT} still suffer from significant performance degradation in CD-FSOD, as shown in Fig. \ref{fig:intro}(c).

Due to the domain gap and limited training data, feature confusion becomes the reason behind the performance degradation of FSOD approaches in CD-FSOD \cite{CD-ViTO}. Such confusion can be divided into object-background confusion and object-object confusion. The object-background confusion refers to the confusion between expected objects and background. As illustrated in Fig. \ref{fig:intro}(a), in underwater scenes, the boundaries between the target object and the background are often ambiguous, leading to missed detections. The object-object confusion refers to the confusion between different classes of objects. As illustrated in Fig. \ref{fig:intro}(b), the similarity between different classes results in false detections. 

In the field of CD-FSOD, CD-ViTO\cite{CD-ViTO} represents the state-of-the-art work, which devises various fine-tuning modules and achieves significant performance improvements. To address object-background confusion, CD-ViTO re-weights manually selected background features and combines them with object features in a weighted sum. However, manually designed features lack adaptability when the target domain distribution differs \cite{CDAL,DA_Faster_R-CNN}. To address object-object confusion, CD-ViTO \cite{CD-ViTO} enhances class distinction by directly adjusting the support class features. However, direct feature adjustment compromises semantic information, affecting the semantic consistency between the query and support feature spaces \cite{SB}. Consequently, CD-ViTO fails to achieve significant improvements in scenarios with severe object-background and object-object confusion, \emph{i.e.}, the NEU-DET\cite{NEUDET} and UODD\cite{UODD} datasets, as shown in Fig. \ref{fig:intro}(c).

To cope with the mentioned issues, we propose a cross-domain few-shot object detection transformer in this work, named CDFormer, to address feature confusion. To address object-background confusion, we propose an object-background distinguishing (OBD) module. This module incorporates a learnable background token to explicitly represent background features. It enhances the distinction between target features and background features by refining their respective representations. To address object-object confusion, we devise an object-object distinguishing (OOD) module to enhance differences between objects of different classes. In summary, the main contributions of this work are as follows:

\begin{itemize}
  \item We propose the CDFormer, a cross-domain few-shot object detection transformer, to deal with the common feature confusion problem, which outperforms state-of-the-art approaches by a large margin.
  \item We devise an object-background distinguishing (OBD) module incorporating a learnable background token, with which we can enhance both object and background perceptions.
  \item We design an object-object distinguishing (OOD) module, which enhances the distinction between different classes by computing the InfoNCE loss between features of object categories and a set of learnable features.

\end{itemize}

\section{Related Work}
\textbf{Few-Shot Object Detection.} Current FSOD approaches can be broadly divided into the one-stage ones and two-stage ones, based on whether they utilize region proposal networks (RPNs). The two-stage approaches, such as \cite{DE-ViT, Meta-RCNN, Retentive_RCNN, DeFRCN}, typically involve the \emph{feature extraction}, \emph{feature interaction}, \emph{RPN}, \emph{similarity computation}, and \emph{detection} steps. These approaches first use the RPN to extract regions of interest from the query image. Next, they compute the similarity between these regions and the support class features. By leveraging region priors, the model achieves better detection accuracy while accommodating diverse support classes. The single-stage approaches, such as \cite{Meta-YOLO, Meta-DETR, FS-DETR}, consist of the \emph{feature extraction}, \emph{feature interaction}, and \emph{detection}, and they exclude the use of region proposals. While the single-stage approaches avoid potential errors from inaccurate proposals, they have not yet demonstrated superior performance compared to two-stage approaches. Moreover, when the number of classes in the test set is unknown, these methods are often designed to process only one support class per feed-forward step. This limits their flexibility in multi-class scenarios. In comparison, the two-stage approaches benefit from more precise region localization and adaptability, while the single-stage approaches emphasize simplicity but face challenges in complex detection scenes.

\textbf{Cross-Domain Few-Shot Object Detection.} The current configuration of CD-FSOD tasks is defined by few-shot, cross-domain conditions with distinct classes in the source and target domains. Under this configuration, MoF-SOD\cite{MoF-SOD} analyzes the impact of different network architectures and pre-trained datasets on performance, and Distill-CDFSOD\cite{Distill_cdfsod} introduces several datasets and a distillation-based approach. CD-ViTO\cite{CD-ViTO} introduces a comprehensive benchmark that spans six target datasets with significant domain style disparities, providing a valuable resource for evaluating CD-FSOD methods across different target domains. To analyze domain gaps, CD-ViTO investigates metrics such as style differences, inter-class variance, and indefinable boundaries. Building on the two-stage DE-ViT architecture \cite{DE-ViT}, CD-ViTO highlights its significant performance degradation in the CD-FSOD task and proposes fine-tuning techniques to address feature confusion issues. Specifically, it tackles object-background confusion and object-object confusion by reweighting manually designed background features combined with object features and directly adjusting object features.

\textbf{Effects of Region Proposal Network (RPN).} FSOD and CD-FSOD approaches typically consist of two components: a detection backbone for feature extraction and a detection head for classification and regression. In this context, the RPN in two-stage approaches serves as an essential part of the detection backbone. On the other hand, single-stage approaches map support classes to class-agnostic spaces, which similarly integrate into the backbone. Both mechanisms contribute to mitigating overfitting on base classes. However, they do not fully address the class and domain bias in the detection head \cite{DA-PRO}. In two-stage detection frameworks, RPN is widely treated as class-agnostic, but is often biased toward seen domains\cite{Retentive_RCNN,OV-DETR}. CD-ViTO \cite{CD-ViTO} explores k-shot fine-tuning of the RPN on cross-domain datasets, but the improvement is limited. This highlights the challenges of using region-based priors. Single-stage frameworks, on the contrary, avoid reliance on RPN-generated proposals by leveraging image-level structures. However, the works like \cite{Meta-YOLO,Meta-DETR} require fine-tuning for new classes, while \cite{FS-DETR} without fine-tuning relies on extensive pre-training on ImageNet-100 to address the lack of inductive bias in the detection head \cite{16x16_Words}.

In contrast, we build a single-stage network by redefining the detection head and design the OBD module around a learnable background token to tackle the object-background confusion issue, while proposing that the OOD module uses contrastive learning to address the object-object confusion problem.

\begin{figure*}[!htb]
\vspace{-0.5em}
  \centering
  \includegraphics[width=0.9\linewidth]{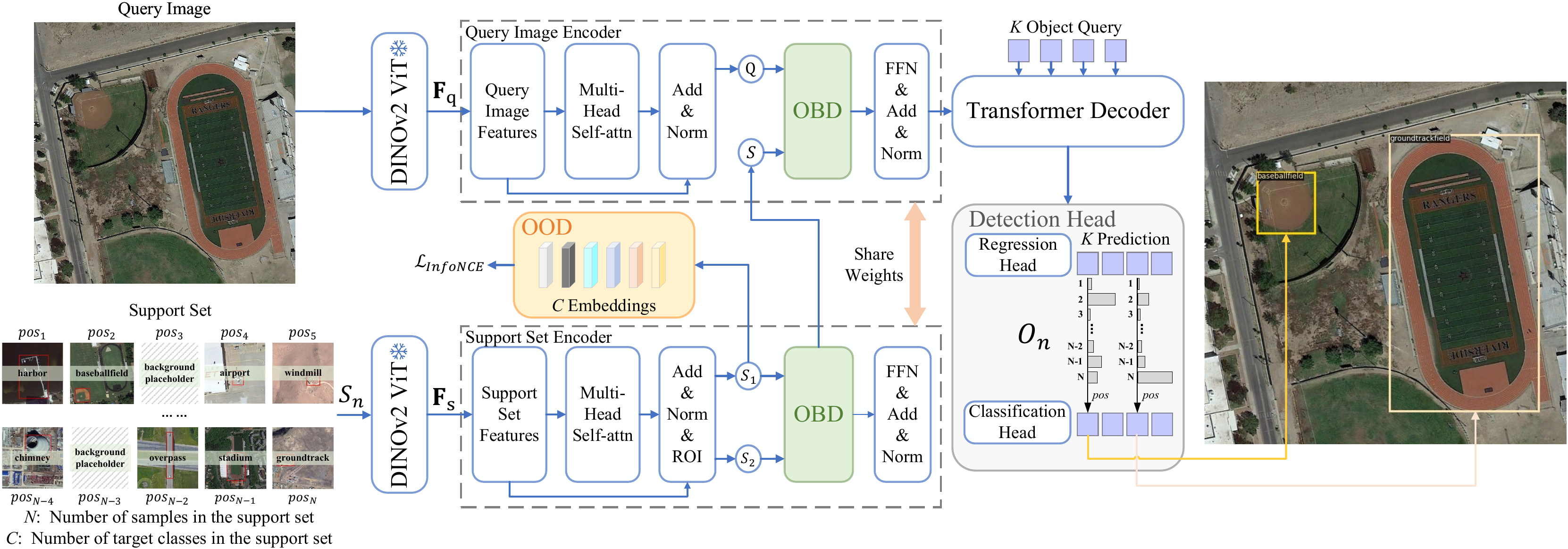}
   \caption{
The overall architecture of our CDFormer. The core of CDFormer consists of the object-background distinguishing (OBD) module and the object-object distinguishing (OOD) module. The term \emph{target classes} refers to the object categories from the dataset represented in the support set.}
   \label{fig:overall}
   \vspace{-0.5em}
\end{figure*}

\begin{figure}[!htb]
  \centering
  \includegraphics[width=\linewidth]{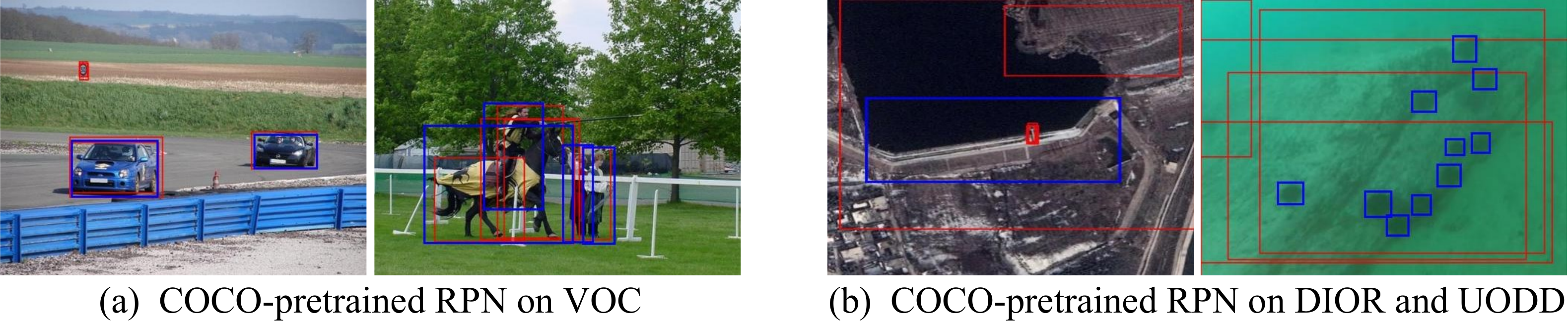}
   \caption{Top 5 results of the COCO-pretrained RPN on same-domain and cross-domain datasets. \textcolor{red}{Red} boxes indicate RPN predictions, and \textcolor{blue}{blue} boxes indicate ground truth. (a) RPN performs well on the same-domain VOC dataset. (b) RPN exhibits significant performance degradation on cross-domain datasets.
}
   \label{fig:RPN}
   \vspace{-1.5em}
\end{figure}

\section{Method}
\subsection{Overview}
Fig. \ref{fig:overall} illustrates the overall architecture of our CDFormer, where we build our baseline by referencing the transformer encoder-decoder structure from DETR \cite{DETR} and Deformable-DETR \cite{deformable-DETR}. The core design of CDFormer addresses two critical challenges in CD-FSOD: object-background confusion and object-object confusion. For the former, we devise an object-background distinguishing (OBD) module to emphasize object features. For the latter, we develop an object-object distinguishing (OOD) module to enhance the difference between target classes.

During the feed-forward of the network, the query image and target classes first pass through the feature extractor to obtain the query image features $\textbf{F}_\text{q}$ and the support class features $\textbf{F}_\text{s}$. These features are then processed using multi-head attention to adjust the query image features and the intrinsic features of each target class. Then, we compute the InfoNCE loss using the adjusted features of the target classes to increase the distance among them, as part of the OOD module. The OBD module addresses object-background confusion by performing three key operations. In the support set branch, it enhances the intrinsic features of each target class while suppressing background interference, ensuring robust class representations. In the query image branch, it further highlights the target class features within the query image features, enabling more accurate alignment with the support set. Additionally, the module explicitly supervises the background to refine the representation of the learnable background token, ensuring a clear separation between object and background features.

\begin{figure}[!htb]
  \centering
  \includegraphics[width=0.9\linewidth]{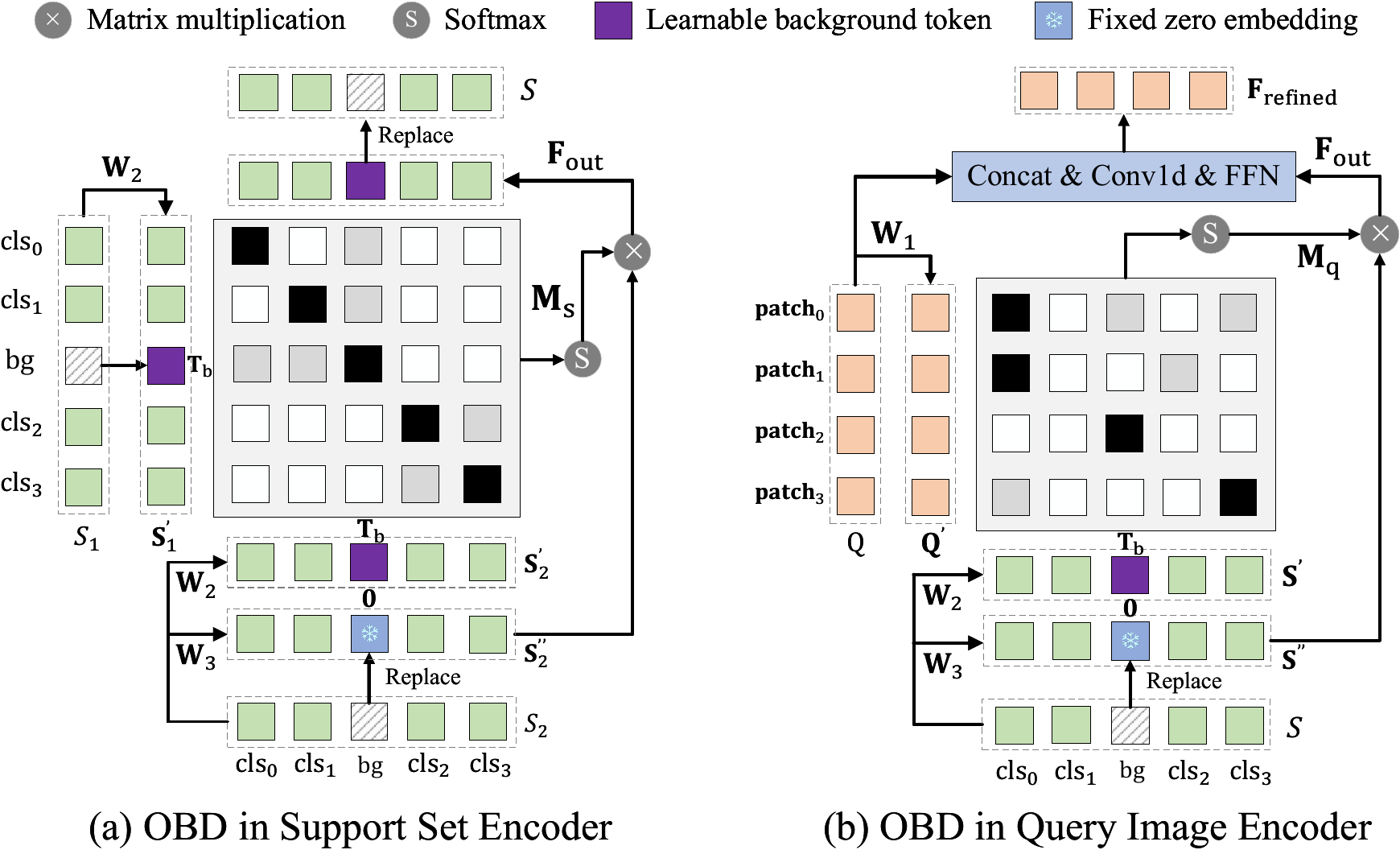}
   \caption{Object feature enhancement (OFE) unit. The OFE serves different purposes in the query image encoder branch and the support set encoder branch. 
}
   \label{fig:OAC}
   \vspace{-1.5em}
\end{figure}

\textbf{Detection Head.} In CD-FSOD, domain bias in feature representations causes the RPN in two-stage approaches to generate numerous erroneous region proposals, as shown in Fig. \ref{fig:RPN}, significantly degrading proposal quality. In contrast, we note that single-stage networks can inherently transform meta-learning methods into class-agnostic matching tasks at corresponding positions. We reinterpret and redefine the detection head for single-stage methods and introduce background placeholders to support inference with an arbitrary number of target classes when the number of classes is unknown. 

Assuming that the support set contains \(N\) samples and the target domain dataset contains \(C\) classes, we use background placeholders to complete the unfilled parts of the support set sequence \({S}_{n}\) with \(n \in \{\mathit{pos}_1, \ldots, \mathit{pos}_N\}\), when $C < N$. In the classification head, we apply a sigmoid activation function to the class prediction embedding vector of each query object, predicting the class probabilities ${O}_{n}$ for $N$ support set sequence positions. By redefining the output sequence format of the classification head, we map the single-stage network to a meta-learning task, which corresponds to position-specific class probabilities, \emph{i.e.}, $\{S_n \xrightarrow{} O_n\} = \{ S_{pos_1} \rightarrow O_{pos_1}, \ldots, S_{pos_N} \rightarrow O_{pos_N} \}$. We achieve class-agnostic meta-learning by reinterpreting and redefining the detection head, alleviating domain bias and enhancing the domain robustness of the network.

\begin{figure}[!htb]
  \centering
  \includegraphics[width=0.85\linewidth]{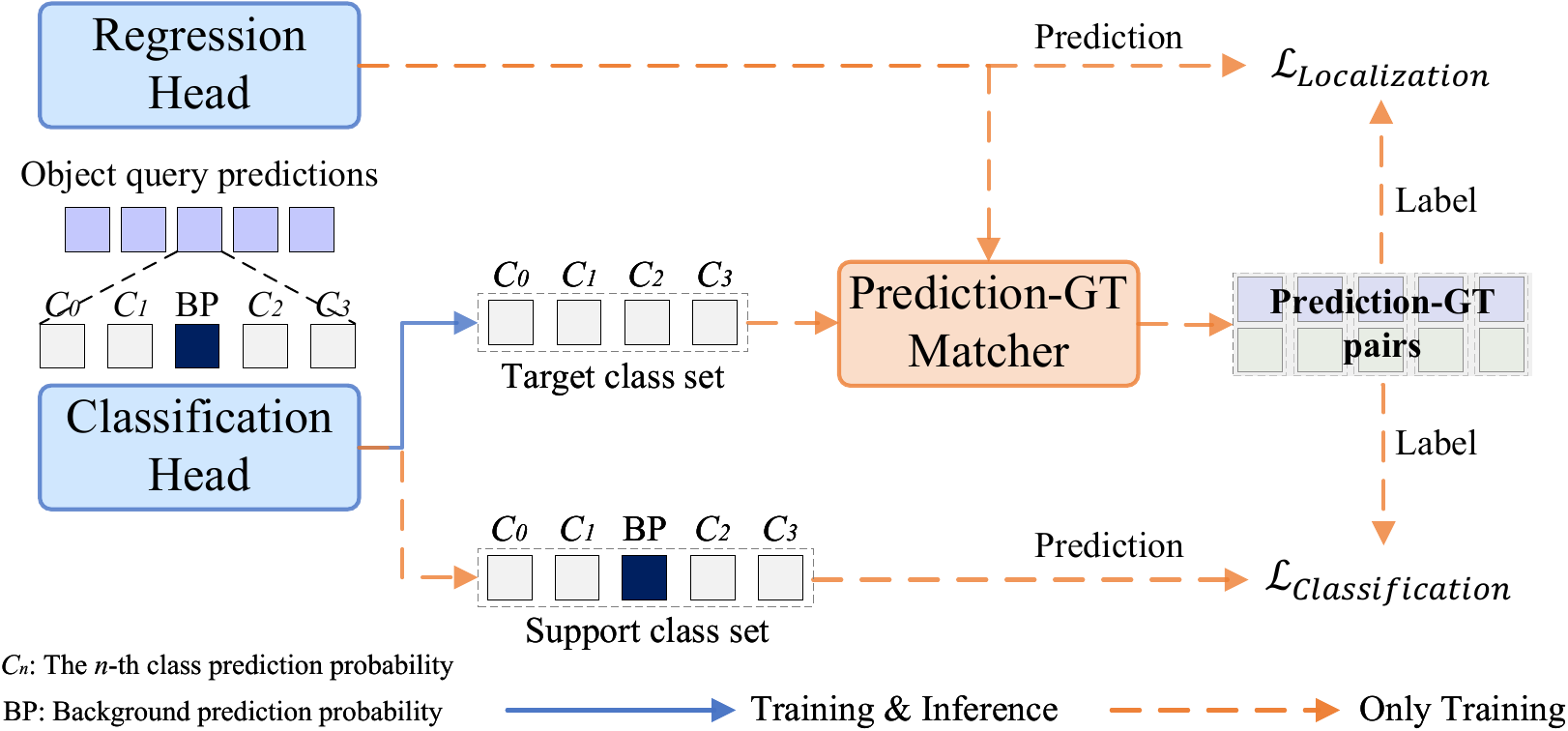}
   \caption{Background feature learning (BFL) unit. As an example, the figure depicts a support set with four target classes and one background placeholder. 
}
   \label{fig:BADH}
   \vspace{-1.5em}
\end{figure}

\begin{figure*}[!htb]
  \centering
  \includegraphics[width=0.9\linewidth]{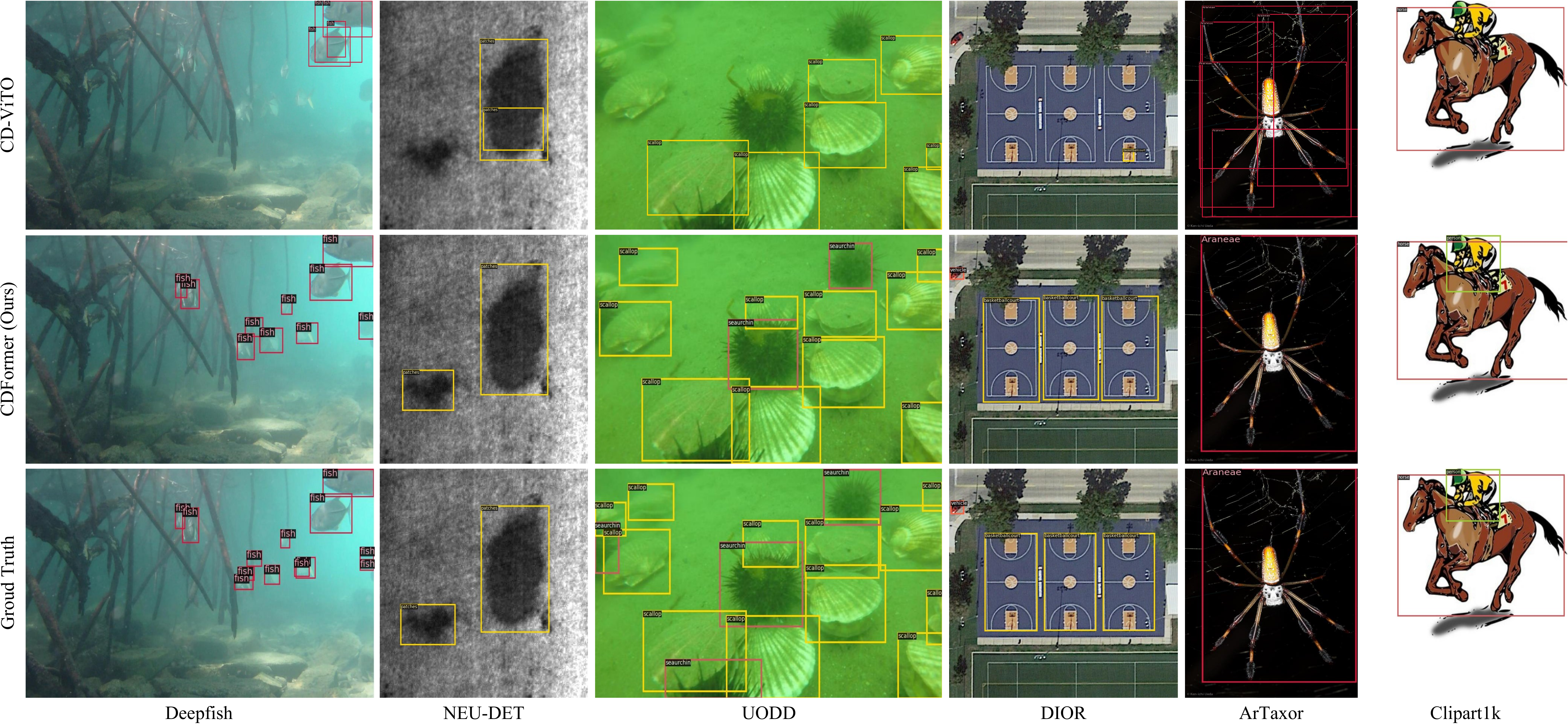}
   \caption{Visualization of the object detection results produced by CD-ViTO\cite{CD-ViTO} and our CDFormer on different datasets.}
   \label{fig:vision}
\end{figure*}

\subsection{Object-Background Distinguishing (OBD) Module}
We introduce a learnable background token $\mathbf{T}_{\text{b}}$ to explicitly represent distinctive background features and design two units around it: object feature enhancement (OFE) and background feature learning (BFL) units. The OFE employs $\mathbf{T}_{\text{b}}$ to separately enhance the distinction between target and background features in the query image branch and the support set branch. Meanwhile, the BFL provides a supervision signal for the background to utilize $\mathbf{T}_{\text{b}}$.

\textbf{Object Feature Enhancement.} The illustration of the OBD is shown in Fig. \ref{fig:OAC}. In the support set encoder branch, the architecture of the OBD is illustrated in Fig. \ref{fig:OAC}(a). For the sake of explanation, we assume that the network input includes a support set of \(N=5\) samples. These samples correspond to \(C=4\) target classes, with one additional background placeholder. The two identical sequences input into the OBD are defined as $ S_1 = S_2 = \{\mathtt{cls}_0, \mathtt{cls}_1, \mathtt{[bg]}, \mathtt{cls}_2, \mathtt{cls}_3\} $, where each positional identifier $\mathtt{cls}_n$ corresponds to a real feature vector $\mathbf{c}_n$, representing the target class at the position of $\mathtt{cls}_n$. Each feature vector $\mathbf{c}_n$ ($n = 0, 1, 2, 3$) is first independently updated through a weight matrix $\mathbf{w}_2$ as $\mathbf{c}_n' = \mathbf{w}_2 \mathbf{c}_n$. The background placeholder $\mathtt{[bg]}$ is excluded from the computation, as it will be replaced later with a learnable background token $\textbf{T}_{\text{b}}$. After all vectors are computed, they are arranged back into their original positions and the background placeholder $\mathtt{[bg]}$ is replaced with the learnable token $\textbf{T}_{\text{b}}$, resulting in $\mathbf{S}_1' = \mathbf{S}_2' = [\mathbf{c}_0', \mathbf{c}_1', \textbf{T}_{\text{b}}, \mathbf{c}_2', \mathbf{c}_3']$, where the positional correspondence of all elements in the original sequences $S_1$ and $S_2$ is preserved. Each feature vector $\mathbf{c}_n$ ($n = 0, 1, 2, 3$) is further updated through a weight matrix $\mathbf{w}_3$ as $\mathbf{c}_n'' = \mathbf{w}_3 \mathbf{c}_n$, while the $\mathtt{[bg]}$ is replaced with a fixed zero vector, resulting in the sequence $\mathbf{S}_2'' = [\mathbf{c}_0'', \mathbf{c}_1'', \mathbf{0}, \mathbf{c}_2'', \mathbf{c}_3'']$. The similarity between $\mathbf{S}_1'$ and $\mathbf{S}_2'$ is calculated using scaled dot product attention\cite{attention_is_all_you_need} as $\mathbf{M}_\text{s} = \text{softmax}((\mathbf{S}_1'(\mathbf{S}_2')^\top)/\sqrt{d})$, where $d$ is the dimension of the feature vectors. The final output feature is extracted by performing matrix multiplication between $\mathbf{M}_\text{s}$ and $\mathbf{S}_2''$, resulting in $\mathbf{F}_{\text{out}} = \mathbf{M}_\text{s} \mathbf{S}_2''$. The similarity matrix $\mathbf{M}_\text{s}$ represents the correlation of each class with the other classes and the background. By performing a matrix multiplication with $\mathbf{S}_2''$, which uses fixed zero vectors to represent the background, we can highlight the features of each class. This treatment mitigates the influence of background features.

In the query image encoder branch, the architecture of OBD is illustrated in Fig. \ref{fig:OAC}(b). The OBD in this branch takes two inputs: the support set $S$ modified by the support set encoder branch, and a set of query image patch features $\mathbf{Q}$. Each query image patch feature in $\mathbf{Q}$ is first transformed using the weight matrix $\mathbf{w}_1$ as $\mathbf{Q}' = \mathbf{w}_1 \mathbf{Q}$. The processing of $S$ follows the same procedure as in the support set encoder branch, resulting in the sequences $\mathbf{S}'$ and $\mathbf{S}''$, where the background in $\mathbf{S}'$ is replaced with a learnable token $\textbf{T}_{\text{b}}$, and the background in $\mathbf{S}''$ is replaced with a fixed zero vector. The similarity between $\mathbf{Q}'$ and $\mathbf{S}'$ is computed as $\mathbf{M}_\text{q} = \text{softmax}((\mathbf{Q}'(\mathbf{S}')^\top)/\sqrt{d})$, where $d$ is the dimension of the feature vectors. During training, the $\textbf{T}_{\text{b}}$ gradually learns to increase its similarity to background features while reducing its similarity to object features through the similarity computation process. To extract class-specific features for each query image patch feature, we perform a matrix multiplication as $\mathbf{F}_{\text{out}} = \mathbf{M}_\text{q} \mathbf{S}''$, which enhances target perception by associating each query image patch feature with the features of the target classes while suppressing background interference. To retain the global features of the query image, the refined representation is computed as $\mathbf{F}_{\text{refined}} = \mathtt{FFN}(\mathtt{Conv1D}(\mathtt{concat}(\mathbf{Q}, \mathbf{F}_{\text{out}})))$, where $\mathtt{concat}(\mathbf{Q}, \mathbf{F}_{\text{out}})$ represents the concatenation of the global features $\mathbf{Q}$ and the extracted features $\mathbf{F}_{\text{out}}$. The $\mathtt{Conv1D}(\cdot)$ operator adjusts the channel dimensions, and the $\mathtt{FFN}(\cdot)$ applies non-linear transformations to refine the feature representation.

\textbf{Background Feature Learning.} This unit is designed to explicitly perceive the background and effectively refine the representation of the learnable background token $\mathbf{T}_{\text{b}}$. As illustrated in Fig. \ref{fig:BADH}, when the background placeholder is presented in the support set, each object query prediction outputs the probabilities that the corresponding position belongs to the background. We constrain these probabilities to guide the explicit aggregation of background representations. Since the background is not a target class, we exclude it from the support set during bipartite matching and use only the predictions of the target classes for matching. We perform the supervision by classification loss and localization loss, as described in \cite{DETR}, where the classification probabilities include the background prediction.

\subsection{Object-Object Distinguishing (OOD) Module}

To allow the network to adjust the features of different target classes, we construct a feature space $\mathbf{T}$ containing $C$ embedding vectors for $C$ target classes, defined as $\mathbf{T} = \{ \mathbf{t}_{1}, \mathbf{t}_{2}, \ldots, \mathbf{t}_{C}\} \in \mathbb{R}^{C \times d}$. Each vector $\mathbf{t}_{i} \in \mathbb{R}^d$ represents the embedding for the $i$-th target class, where $i = 1, 2, \ldots, C$. Additionally, let $\mathbf{F}$ denote the features generated by the support set encoder branch, where the target class features extracted are denoted as $\mathbf{f}_{i}$. Thus, $\mathbf{F} = \{ \mathbf{f}_{1}, \mathbf{f}_{2}, \ldots, \mathbf{f}_{C} \} \in \mathbb{R}^{C \times d}$, with each vector $\mathbf{f}_{i} \in \mathbb{R}^d$. We apply the InfoNCE loss to optimize the feature alignment between the feature vectors in $\mathbf{F}$ and the learnable embedding vectors in $\mathbf{T}$:

\begin{align}
\mathcal{L}_\text{OOD} = -\frac{1}{C} \sum_{i=1}^{C} \left( \log \frac{ \exp(\mathbf{f}^\top_{i} \mathbf{t}_{i} / \tau)}{ \sum_{j=1}^{C} \exp(\mathbf{f}^\top_{i} \mathbf{t}_{j} / \tau)} \right)
\end{align}

\noindent where $\tau$ denotes the temperature hyperparameter. The feature space $\mathbf{T}$ is initialized randomly following a normal distribution. Using the InfoNCE loss, we maximize the similarity between each target class feature and its corresponding feature in the space \(\mathbf{T}\), while minimizing its similarity to features of other classes. By increasing the inter-class distance, we enhance the distinction between objects of different classes.

\begin{table}[h!]
    \vspace{-1em}
    \centering
    \caption{10-shot ablation experiment with fine-tuning.}
    \large
    \renewcommand{\arraystretch}{1.0} 
    \resizebox{\columnwidth}{!}{ 
    \begin{tabular}{@{}lcc|ccccccc@{}}
        \toprule
        \textbf{Baseline} & \textbf{OBD} & \textbf{OOD} & \textbf{ArTaxOr} & \textbf{Clipart} & \textbf{DIOR} & \textbf{Deepfish} & \textbf{NEU-DET} & \textbf{UODD} & \textbf{Avg.} \\ 
        \midrule
        \checkmark & & & 58.5 & 53.6 & 27.4 & 33.0 & 13.9 & 19.8 & 34.3 \\
        \checkmark & \checkmark & & 67.1 & 58.1 & 29.3 & 34.0 & 16.5 & 22.7 & 37.9 \\
        \checkmark & \checkmark & \checkmark & 68.7 & 59.0 & 32.5 & 35.5 & 18.1 & 26.4 & 40.0 \\ 
        \bottomrule
    \end{tabular}
    }
\label{with}
\end{table}


\begin{table}[h!]
    \vspace{-1em}
    \centering
    \caption{10-shot ablation experiment without fine-tuning.}
    \large
    \renewcommand{\arraystretch}{1.0} 
    \resizebox{\columnwidth}{!}{ 
    \begin{tabular}{@{}lcc|ccccccc@{}}
        \toprule
        \textbf{Baseline} & \textbf{OBD} & \textbf{OOD} & \textbf{ArTaxOr} & \textbf{Clipart} & \textbf{DIOR} & \textbf{Deepfish} & \textbf{NEU-DET} & \textbf{UODD} & \textbf{Avg.} \\ 
        \midrule
        \checkmark & & & 20.8 & 51.1 & 6.8 & 27.0 & 0.7 & 10.2 & 19.4 \\
        \checkmark & \checkmark & & 40.9 & 53.2 & 7.0 & 25.6 & 1.8 & 12.0 & 23.4 \\
        \checkmark & \checkmark & \checkmark & 37.3 & 53.5 & 7.9 & 25.7 & 4.0 & 16.7 & 24.2 \\ 
        \bottomrule
    \end{tabular}
    }
\label{without}
\end{table}

\section{Experiments}
\subsection{Experimental Setup}
Our model is pretrained on the COCO\cite{COCO} dataset and then fine-tuned and evaluated on six other cross-domain datasets using k-shot learning. The six datasets are ArTaxOr\cite{ArTaxOr}, Clipart1k\cite{Clipart1k}, DIOR\cite{DIOR}, DeepFish\cite{DeepFish}, NEU-DET\cite{NEUDET}, and UODD\cite{UODD}.

Following CD-ViTO\cite{CD-ViTO}, we primarily evaluate our method under fine-tuning conditions to ensure a fair comparison with previous approaches. We report the mean average precision at IoU thresholds from 0.5 to 0.95 (mAP@0.5:0.95) with or without fine-tuning as the evaluation metric. We present results for 1/5/10 shots across six cross-domain datasets under the both settings.

\begin{table}[h!t] 
    \centering
    \caption{The mAP values of different approaches produced on the CD-FSOD benchmarks in case of 1-shot, 5-shot, and 10-shot. Here ``w/ FT'' and ``w/o FT'' denote the results produced with and without fine-tuning, respectively.}
    \scriptsize
    \setlength{\tabcolsep}{0.45pt} 

    \resizebox{0.48\textwidth}{!}{  
    \begin{tabular}{@{}lcccccccccc@{}}
        \toprule
        Method & Backbone & ArTaxOr & Clipart1k & DIOR & DeepFish & NEU-DET & UODD & Avg. \\ \midrule
        \textbf{1-SHOT} & & & & & & & & \\ 
        Meta-RCNN\cite{Meta-RCNN} & ResNet50 & 2.8 & - & 7.8 & - & - & 3.6 & / \\
        TFA w/cos\cite{TFA} & ResNet50 & 3.1 & - & 8.0 & - & - & 4.4 & / \\
        FSCE\cite{FSCE} & ResNet50 & 3.7 & - & 8.6 & - & - & 3.9 & / \\
        DeFRCN\cite{DeFRCN} & ResNet50 & 3.6 & - & 9.3 & - & - & 4.5 & / \\
        Distill-CDFSOD\cite{Distill_cdfsod} & ResNet50 & 5.1 & 7.6 & 10.5 & / & / & 5.9 & / \\
        ViTDeT-FT\cite{ViTDeT} & ViT-B/14 & 5.9 & 6.1 & 12.9 & 0.9 & 2.4 & 4.0 & 5.4 \\
        Detic\cite{Detic} & ViT-L/14 & 0.6 & 11.4 & 0.1 & 0.9 & 0.0 & 0.0 & 2.2 \\
        Detic-FT\cite{Detic} & ViT-L/14 & 3.2 & 15.1 & 4.1 & 9.0 & \textcolor{blue}{3.8} & 4.2 & 6.6 \\
        DE-ViT\cite{DE-ViT} & ViT-L/14 & 0.4 & 0.5 & 2.7 & 0.4 & 0.4 & 1.5 & 1.0 \\  
        DE-ViT-FT\cite{DE-ViT} & ViT-L/14 & 10.5 & 13.0 & 14.7 & 19.3 & 0.6 & 2.4 & 10.1 \\
        CD-ViTO\cite{CD-ViTO} & ViT-L/14 & {21.0} & {17.7} & \textcolor{red}{17.8} & {20.3} & 3.6 & 3.1 & {13.9} \\ 
        \rowcolor{gray!30}  
        {CDFormer w/o FT (ours)} & ViT-L/14 & \textcolor{blue}{29.4} & \textcolor{blue}{49.6} & {8.0} & \textcolor{blue}{27.7} & 3.0 & \textcolor{blue}{12.2} & \textcolor{blue}{21.6} \\ 
        \rowcolor{gray!30}  
        {CDFormer w/FT (ours)} & ViT-L/14 & \textcolor{red}{36.0} & \textcolor{red}{54.0} & \textcolor{blue}{16.3} & \textcolor{red}{34.5} & \textcolor{red}{7.4} & \textcolor{red}{12.7} & \textcolor{red}{26.8} \\
        \midrule
        \textbf{5-SHOT} & & & & & & & & \\
        Meta-RCNN\cite{Meta-RCNN} & ResNet50 & 8.5 & - & 17.7 & - & - & 8.8 & / \\
        TFA w/cos\cite{TFA} & ResNet50 & 8.8 & - & 18.1 & - & - & 8.7 & / \\
        FSCE\cite{FSCE} & ResNet50 & 10.2 & - & 18.7 & - & - & 9.6 & / \\
        DeFRCN\cite{DeFRCN} & ResNet50 & 9.9 & - & 18.9 & - & - & 9.9 & / \\
        Distill-CDFSOD\cite{Distill_cdfsod} & ResNet50 & 12.5 & 23.3 & 19.1 & 15.5 & \textcolor{red}{16.0} & 12.2 & 16.4 \\
        ViTDeT-FT\cite{ViTDeT} & ViT-B/14 & 20.9 & 23.3 & 23.3 & 9.0 & 13.5 & 11.1 & 16.9 \\
        Detic\cite{Detic} & ViT-L/14 & 0.6 & 11.4 & 0.1 & 0.9 & 0.0 & 0.0 & 2.2 \\
        Detic-FT\cite{Detic} & ViT-L/14 & 8.7 & 20.2 & 12.1 & 14.3 & 14.1 & 10.4 & 13.3 \\
        DE-ViT\cite{DE-ViT} & ViT-L/14 & 10.1 & 5.5 & 7.8 & 2.5 & 1.5 & 3.1 & 5.1 \\
        DE-ViT-FT\cite{DE-ViT} & ViT-L/14 & 38.0 & 38.1 & 23.4 & 21.2 & 7.8 & 5.0 & 22.3 \\
        CD-ViTO\cite{CD-ViTO} & ViT-L/14 & \textcolor{blue}{47.9} & {41.1} & \textcolor{blue}{26.9} & {22.3} & 11.4 & 6.8 & \textcolor{blue}{26.1} \\ 

        \rowcolor{gray!30}  
        {CDFormer w/o FT (ours)} & ViT-L/14 & {38.4} & \textcolor{blue}{53.6} & {7.9} & \textcolor{blue}{24.6} & 3.6 & \textcolor{blue}{16.5} & {24.1} \\ 
        \rowcolor{gray!30}  
        {CDFormer w/FT (ours)} & ViT-L/14 & \textcolor{red}{65.0} & \textcolor{red}{58.9} & \textcolor{red}{28.1} & \textcolor{red}{31.7} & \textcolor{blue}{15.0} & \textcolor{red}{23.8} & \textcolor{red}{37.1} \\
        \midrule
        \textbf{10-SHOT} & & & & & & & & \\
        Meta-RCNN\cite{Meta-RCNN} & ResNet50 & 14.0 & - & 20.6 & - & - & 11.2 & / \\
        TFA w/cos\cite{TFA} & ResNet50 & 14.8 & - & 20.5 & - & - & 11.8 & / \\
        FSCE\cite{FSCE} & ResNet50 & 15.9 & - & 21.9 & - & - & 12.0 & / \\
        DeFRCN\cite{DeFRCN} & ResNet50 & 15.5 & - & 22.9 & - & - & 12.1 & / \\
        Distill-CDFSOD\cite{Distill_cdfsod} & ResNet50 & 18.1 & 27.3 & 26.5 & 15.5 & \textcolor{red}{21.1} & 14.5 & 20.5 \\
        ViTDeT-FT\cite{ViTDeT} & ViT-B/14 & 23.4 & 25.6 & 29.4 & 6.5 & 15.8 & 15.6 & 19.4 \\
        Detic\cite{Detic} & ViT-L/14 & 0.6 & 11.4 & 0.1 & 0.9 & 0.0 & 0.0 & 2.2 \\
        Detic-FT\cite{Detic} & ViT-L/14 & 12.0 & 22.3 & 15.4 & 17.9 & 16.8 & 14.4 & 16.5 \\
        DE-ViT\cite{DE-ViT} & ViT-L/14 & 9.2 & 11.0 & 8.4 & 2.1 & 1.8 & 3.1 & 5.9 \\
        DE-ViT-FT\cite{DE-ViT} & ViT-L/14 & 49.2 & 40.8 & 25.6 & 21.3 & 8.8 & 5.4 & 25.2 \\
        CD-ViTO\cite{CD-ViTO} & ViT-L/14 & \textcolor{blue}{60.5} & {44.3} & \textcolor{blue}{30.8} & {22.3} & 12.8 & 7.0 & \textcolor{blue}{29.6} \\ 
        
        \rowcolor{gray!30}  
        {CDFormer w/o FT (ours)} & ViT-L/14 & {37.3} & \textcolor{blue}{53.5} & {7.9} & \textcolor{blue}{25.7} & 4.0 & \textcolor{blue}{16.7} & {24.2} \\ 
        \rowcolor{gray!30}  
        {CDFormer w/FT (ours)} & ViT-L/14 & \textcolor{red}{68.7} & \textcolor{red}{59.0} & \textcolor{red}{32.5} & \textcolor{red}{35.5} & \textcolor{blue}{18.1} & \textcolor{red}{26.4} & \textcolor{red}{40.0} \\
        \bottomrule
    \end{tabular}
    }
\label{all}
\vspace{-2em}
\end{table}

\subsection{Ablation Study}
Ablation studies are conducted on the proposed baseline, along with the OBD and OOD modules, to evaluate the contributions of each component. It is important to note that our modules are universal and not specifically tailored for cross-domain fine-tuning scenarios, unlike the modules designed in CD-ViTO\cite{CD-ViTO}. Hence, we perform ablation studies both with and without fine-tuning, with results shown in Table \ref{with} and Table \ref{without}. Additionally, since the OBD module is an essential feature interaction module in our CDFormer, it cannot be entirely removed during the ablation study. To investigate its effectiveness, the ablation for OBD is conducted by setting the support set to a single class, \emph{i.e.}, $N = C = 1$, where the background is not involved in the OBD module.

Table \ref{with} and Table \ref{without} show that our constructed baseline still has a degree of domain robustness, even in the case of without fine-tuning. Moreover, the two proposed modules show significant improvements to address object-background confusion and object-object confusion, particularly on the challenging NEU-DET and UODD datasets.

\subsection{Performance Evaluation}
Table \ref{all} presents a comparison of our 1/5/10-shot mAP metrics on the benchmark against other approaches. It is evident that our method significantly outperforms existing approaches, particularly during fine-tuning, with notable improvements on the NEU-DET and UODD datasets. This is because of the ability of our method that addresses the issues of object-background confusion and object-object confusion. Even without fine-tuning on the target dataset, our method still substantially outperforms the previous ones that do not require fine-tuning \cite{DE-ViT,Detic}. In the 1-shot scenario, our method even surpasses the fine-tuned state-of-the-art method CD-ViTO\cite{CD-ViTO}, demonstrating our domain robustness. 

\begin{figure}[!htb]
  \centering
  \includegraphics[width=0.65\linewidth]{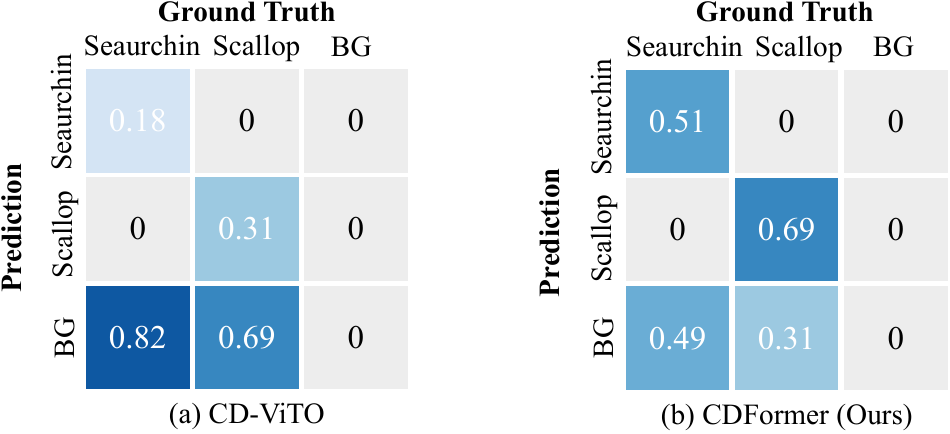}
   \caption{10-shot fine-tuning confusion matrices of CD-ViTO \cite{CD-ViTO} and our CDFormer on the UODD dataset. BG represents the background.}
   \label{fig:matrix}
\end{figure}


The 10-shot fine-tuning class confusion matrix for the UODD dataset in underwater scenes is shown in Fig. \ref{fig:matrix}, demonstrating that our method can better distinguish between object and background. The visualization results in Fig. \ref{fig:vision} demonstrate that our method significantly improves performance on the Deepfish, NEU-DET, and UODD datasets, where object-background confusion and object-object confusion are more pronounced. Moreover, our approach exhibits robust performance across other cross-domain datasets.

\section{Conclusions}
In this work, we have proposed CDFormer, an innovative cross-domain few-shot detection architecture designed to address the feature confusion challenge in CD-FSOD. CDFormer integrates two core modules, \emph{i.e.}, an object-background distinguishing module to mitigate object-background confusion, and an object-object distinguishing module to resolve object-object confusion. Experimental results show that our method achieves state-of-the-art performance, significantly surpassing previous approaches, particularly on datasets with severe feature confusion, such as NEU-DET and UODD.

\bibliographystyle{IEEEbib}
\bibliography{ref}

\end{document}